\pdfoutput=1

\documentclass[11pt]{article}

\usepackage[final]{acl}
\usepackage{algorithm}
\usepackage{algorithmic}
\usepackage{dingbat}

\usepackage{times}
\usepackage{latexsym}
\usepackage{amsmath}
\usepackage{pifont}

\usepackage[T1]{fontenc}

\usepackage[utf8]{inputenc}

\usepackage{microtype}

\usepackage{inconsolata}

\usepackage{amsmath}
\usepackage{amssymb}
\usepackage{amsthm}
\usepackage{txfonts}
\usepackage{microtype}
\usepackage{graphicx}
\usepackage{subcaption}
\usepackage{booktabs} 
\usepackage{multicol}
\usepackage{multirow}
\usepackage{makecell}
\usepackage{pgfplots}
\pgfplotsset{compat=newest}
\usepackage{caption}
\usepackage{enumitem}
\usepackage{pifont}
\usepackage{xcolor}
\usepackage{txfonts}
\usepackage{tabularx}
\usepackage{float}
\usepackage{bbm}
\usepackage{etex}
\usepackage{svg}
\usepackage[table]{xcolor}
\definecolor{lightpurple}{RGB}{245,238,255} 
\definecolor{ourRed}{RGB}{209,41,32} %
\definecolor{ourBlue}{RGB}{38,159,255} %
\definecolor{oursbg}{RGB}{245,238,255}
\definecolor{mygreen}{RGB}{34,139,34}

\title{From Passive Response to Proactive Correction: \\ Enhancing LLM Robustness Against Input Fact Perturbations}

\author{
Ping Wang$^{1}$ \quad
Xiangguo Sun$^{2}$\thanks{\, Correspondence to Xiaofeng Meng and Xiangguo Sun.}, \quad
Bingbing Xu$^{1}$ \quad
Guocong Li$^{3}$ \quad
Xiaofeng Meng$^{1}$\footnotemark[1]\\
$^{1}$Renmin University of China \quad
$^{2}$Southeast University \quad
$^{3}$Zhejiang University\\
\texttt{\{p\_wang,xvbingbing,xfmeng\}@ruc.edu.cn}\\
\texttt{hsiang-kuo.sun@seu.edu.cn, liguocong@zju.edu.cn}
}

\begin{document}
\maketitle

\begin{abstract}
Large language models (LLMs) frequently produce confident yet factually incorrect responses when user inputs contain misleading premises, a phenomenon we attribute to fact perturbations in the input. Existing approaches to hallucination mitigation typically assume reliable user inputs, overlooking how such factual errors can actively mislead model reasoning. To address this vulnerability, we propose DEDUCE, a three-stage framework that transforms LLMs from passive responders into proactive error correctors. 
DEDUCE operates in three stages: (1) detect errors through fine-grained fact extraction and verification; (2) devise correction strategies via multi perspective deliberation; and (3) correct misconceptions while delivering reliable answers. We also present MisFactQA, a dataset containing factual errors of varying degrees, and propose new metrics for evaluating model robustness. Experiments on TruthfulQA, FalseQA, and our MisFactQA benchmark demonstrate that DEDUCE significantly improves both accuracy and error correction capability. Consistent gains across Qwen, LLaMA, and Gemma families confirm its effectiveness and scalability.\footnotemark
\end{abstract}

\footnotetext{\url{https://github.com/WangPing-Kayla/Deduce}}

\begin{figure}[ht]
    \centering
    \includegraphics[width=0.48\textwidth]{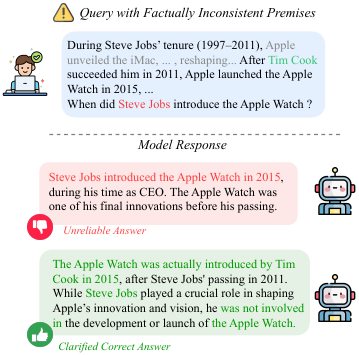}
    \caption{A query containing factually incorrect or inconsistent premises can mislead an LLM into producing a confident but unreliable answer (\textcolor{red}{red}). The desired behavior is to identify the flawed premise, explicitly clarify the error, and then provide a corrected, factual response (\textcolor{mygreen}{green}).}
    \label{fig:Figure_1}
    \vspace{-8pt}
\end{figure}

\vspace{-5pt}
\section{Introduction}
\vspace{-3pt}
In recent years, large language models (LLMs) have demonstrated remarkable capabilities in encoding real-world knowledge within their parameters and using it to assist with complex tasks~\citep{ge2023openagi,lyu2024retrieve,li2025fundamental}. Despite their impressive performance on a wide range of generative tasks, including translation~\citep{lu2024llamax}, mathematical reasoning~\citep{xia2025evaluating}, code generation~\cite{11511832}, and value alignment~\citep{xu-etal-2025-towards}, LLMs remain highly vulnerable to hallucinations, producing outputs that appear plausible yet are factually incorrect~\citep{huang2025survey, li-etal-2025-misleading, li-etal-2026-debate}. 

Existing research typically relieves hallucination along pre-training, supervised fine-tuning, and inference stages, nearly covering the entire pipeline of developing an LLM. Specifically, in the pre-training stage, hallucinations are traced to flawed corpora~\cite{penedo2023refinedweb,li2023textbooks}. During supervised fine-tuning, hallucinations frequently arise when models are compelled to answer beyond their parametric knowledge, prompting work on teaching models to appropriately express uncertainty~\cite{3692070.3692393}. The inference stage has attracted the most attention due to its flexibility and low cost~\cite{ji2023towards, zhang2024self, tonmoy2024comprehensive, cheng-etal-2025-think,sriramanan2024llm, li2026mol}, where hallucinations stem from imperfect decoding and insufficient grounding. Accordingly, prior studies have proposed faithful decoding strategies~\cite{dhuliawala2024chain} , retrieval-augmented generation~\citep{bechard2024reducing}, and uncertainty-based detection methods~\cite{cheng2024integrative}.

Although progress has been made, current research still focuses almost entirely on the product end, that is, on improving LLM capabilities. This overlooks a notable reality at the user end (shown in Figure~\ref{fig:Figure_1}): in real-world applications, users are rarely domain experts, and their queries often contain factual errors. These are factual mistakes stemming from flawed cognition, including incorrect assumptions, contradictory statements, and complex errors. Such errors can  perturb LLM reasoning processes and induce hallucinations as well.

\begin{figure}[t]
    \centering
    \includegraphics[width=0.5\textwidth]{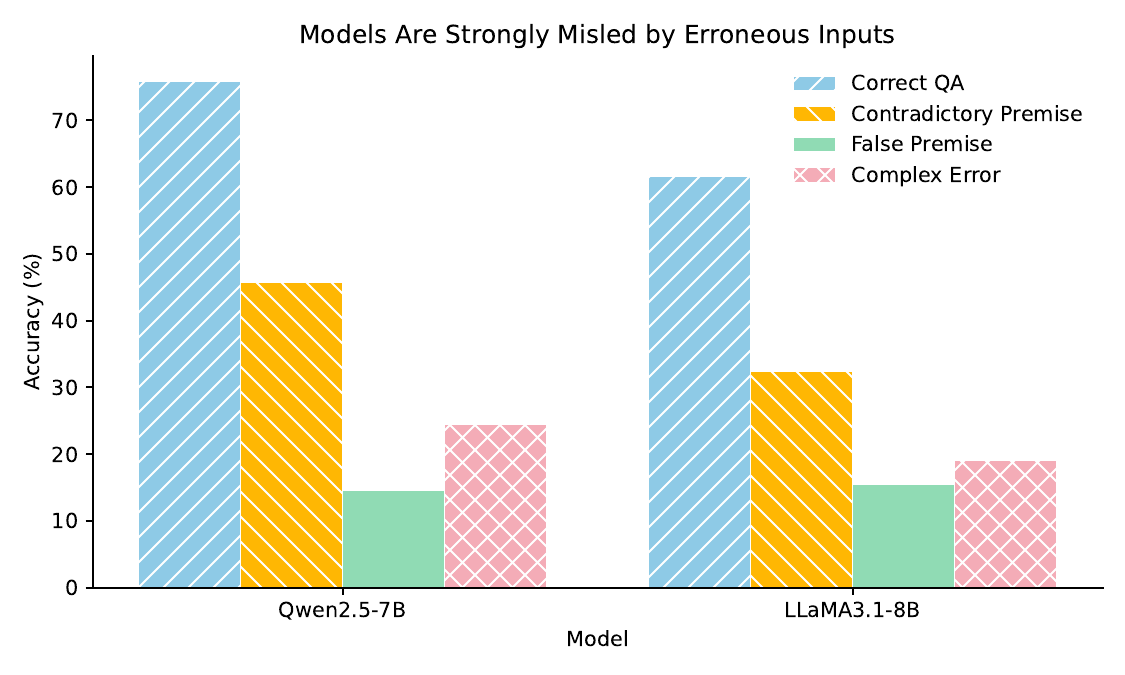}
    \caption{Models are strongly misled by erroneous inputs. On Qwen2.5-7B and LlaMA-3.1-8B, we evaluate four query types built from the same underlying information. Accuracy is high on correct questions (blue) but drops by 30\%-60\% points when errors are injected.}

    \label{fig:misleading performance}
    \vspace{-15pt}
\end{figure}

In this paper, we identify a critical yet frequently neglected factor: \textbf{even if model development is optimized at every stage against hallucinations, the input quality of real-world users can still lead to inevitable hallucination issues in LLMs.} Figure~\ref{fig:misleading performance} further demonstrates this, showing that even injecting a small number of factual errors into the input (e.g., false premises or contradictory descriptions) can lead to a sharp drop in model accuracy. Surprisingly, even when LLMs possess correct underlying knowledge~\citep{yuan-etal-2024-whispers,guo2025protect}, such misleading inputs can override internal representations and derail reasoning. 
Existing methods typically adopt two passive strategies for inputs with false premises or factual errors: they either ignore the errors, thereby propagating misinformation, or attempt self correction but often fail because reasoning is biased by the input.

In light of the above, this work aims to advance research in this direction by addressing the following core question: how can we enhance the reliability of LLM responses when the input contains factual errors? Tackling this problem is critically important yet particularly hard due to three main challenges: \textit{\textbf{Challenge 1}}, user errors are often implicit and varied in form, making them difficult for models to precisely identify and localize. Such errors may be embedded as false statements, incorrect presuppositions, internal inconsistency, or layered in compound queries, requiring deep understanding to uncover. \textit{\textbf{Challenge 2}}, even when errors are detected, it is non-trivial for a model to reliably correct all inaccuracies while still providing a helpful and accurate answer. The correction process itself can introduce new inconsistencies or lead to incomplete responses, demanding both rigorous error resolution and faithful adherence to the corrected context.
\textit{\textbf{Challenge 3}}, existing benchmarks lack complex erroneous queries, and conventional evaluation metrics fail to capture nuanced model behaviors. In real scenarios, models may only partially correct errors, correct but refuse to answer, or answer without rectifying the error.



To address these challenges, we conduct a systematic investigation into input-induced hallucinations and propose a framework that transforms LLMs from passive responders into proactive evaluators. For the \textbf{first challenge}, we introduce a fine-grained \textit{Detect} module that precisely locates and analyzes factual errors in user inputs. For the \textbf{second challenge}, we propose a strategy generation module coupled with a \textit{Multi-Perspective Strategy Deliberation} mechanism, which generates and rigorously verifies response strategies tailored to the diagnosed errors. For the \textbf{third challenge}, we offer a dataset that simulates complex real-world user factual errors, and introduce new evaluation metrics to assess how well models respond when confronted with misleading information. Our contributions are summarized as follows:
\begin{itemize}
\vspace{-2pt}
  \setlength{\itemsep}{2pt}
  \setlength{\parskip}{0pt}
  \setlength{\topsep}{2pt}
\item We present a noteworthy perspective on mitigating hallucinations in LLMs, highlighting that they arise not only by model fabrication but also from factual perturbations in user inputs that mislead model reasoning.
 \item We propose a ``detect, devise and correct'' framework to enhance models' robustness and reliability under misleading input. 
 \item We offer MisFactQA, a more comprehensive and objective benchmarking dataset containing diverse 
and complex factual errors, with fine-grained metrics 
to evaluate model robustness against misleading inputs.\end{itemize}

\begin{figure*}[t]
    \centering
    \includegraphics[width=1\textwidth]{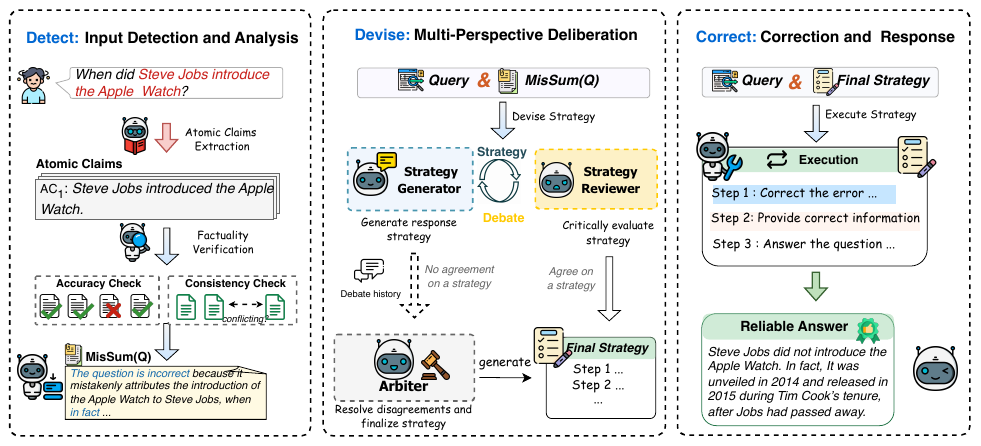}
    \vspace{-17pt}
    \caption{Overall architecture of our DEtect-Devise-CorrEct Framework.}
    \vspace{-12pt}
    \label{fig:main}
\end{figure*}

\section{Background and Problem Statement}
\vspace{-5pt}
\paragraph{Motivation} As previously mentioned, users are rarely domain experts, and their queries often contain factual errors that can perturb LLM reasoning. We argue that robust LLMs should enhance the capacity of
identifying and correcting such errors, rather than passively inherit flawed inputs.

\paragraph{Problem Definition}

Let $Q=(q,\mathcal{K})$ denote a user input, where $q$ is the query and $\mathcal{K}=\{k_1,\dots,k_m\}$ is the set of factual units extracted from it, encompassing both explicit statements and implicit assumptions. We define an indicator function $\mathcal{E}(Q)\in\{0,1\}$, where $\mathcal{E}(Q)=1$ identifies $Q$ as a \textit{factually perturbed input}. Such perturbations take three forms: false premise, where some $k_i$ is factually incorrect and causes the model to reason from a wrong starting point; factual contradiction, where inconsistent elements co-exist in $\mathcal{K}$, leaving the model without a reliable basis for reasoning; and compound error, where multiple erroneous units co-occur such that detecting one does not prevent the remaining errors from misleading the model. When $\mathcal{E}(Q)=1$, LLMs are prone to distorted intermediate reasoning trajectories, ultimately producing responses that inherit or amplify misinformation.


\vspace{-5pt}
\paragraph{Objective}
We formally define the task of Fact-Perturbed Question Answering (\textbf{FPQA}), aiming to enhance language model robustness against misleading inputs. Specifically, we develop a framework on top of various LLM backbones $\mathcal{M}^*$, which is capable of: (1) accurately detecting when $\mathcal{E}(Q) = 1$, (2) explicitly correcting factual errors in $\mathcal{K}$, and (3) generating reliable responses grounded in verified knowledge rather than erroneous user context.

\section{DEDUCE: \textbf{DE}tect-\textbf{D}evise-\textbf{C}orr\textbf{E}ct}

In this section, we present our input-side hallucination mitigation framework, \textsc{DEDUCE}, which addresses a fundamental limitation of existing approaches: when facing factually perturbed inputs, models either propagate misinformation by answering directly, or fail at self-correction by remaining trapped within their own reasoning biases. The core insight of \textsc{DEDUCE} is to reformulate this challenge into an explainable, debatable, and verifiable collaborative process. As illustrated in Figure~\ref{fig:main}, \textsc{DEDUCE} comprises three synergistic modules, namely Detect, Devise, and Correct, that jointly enable the model to proactively surface errors, strategically reason under misleading inputs, and produce reliable, verifiably grounded responses.

\subsection{Detect: \textbf{Input Detection and Analysis}}

Errors in user queries are often concealed within narratives that appear plausible. When a model perceives the input as a whole, it is susceptible to surface level semantic misleading and prone to overlooking subtle but critical inaccuracies. To address this, we introduce an \textbf{atomic fact detection mechanism} that decomposes $Q$ into a set of minimal, independently verifiable factual units.

\paragraph{Atomic Fact Decomposition}

Given a query \(Q\), we define a decomposition function \(\phi\) that maps \(Q\) to its set of atomic facts:
{\setlength{\abovedisplayskip}{4pt}
\setlength{\belowdisplayskip}{4pt}
\setlength{\abovedisplayshortskip}{2pt}
\setlength{\belowdisplayshortskip}{2pt}
\begin{equation}
\phi(Q) = \{AC_1, AC_2, \dots, AC_n\},
\label{eq:decomposition}
\end{equation}
}
where each \(AC_i\) is a minimal, self-contained factual assertion extracted from the query.

 First, it \textit{surfaces implicit errors}: even when the query appears superficially coherent, all latent factual errors are exposed, enabling more comprehensive detection of potential misinformation. Second, it provides an \textit{actionable basis for verification}: each unit is independently assessed via a truthfulness check $\mathcal{F}(\cdot)$ and a pairwise consistency check $\mathcal{C}(\cdot,\cdot)$, defined as:

\vspace{-7pt}
 
\begin{equation}
    \mathcal{F}(AC_i) =
    \begin{cases}
        1, & \text{if } AC_i \text{ is factually incorrect}, \\
        0, & \text{otherwise}.
    \end{cases}
    \label{eq:factcheck}
\end{equation}
\vspace{-6pt}
\begin{equation}
\mathcal{C}(AC_i, AC_j) =
\begin{cases}
1, & \text{if } AC_i \text{ and } AC_j \text{ are} \\
   & \text{mutually inconsistent}, \\
0, & \text{otherwise}.
\end{cases}
\end{equation}

The two checks give rise to disjoint error sets $\mathcal{E}_{\text{fact}} = \{AC_i \mid \mathcal{F}(AC_i)=1\}$ and $\mathcal{E}_{\text{conflict}} = \{(AC_i, AC_j) \mid \mathcal{C}(AC_i,AC_j)=1\}$, enabling precise identification of both error type and location. The overall perturbation indicator defined in Eq.~\ref{eq:eq_indicator} can thereby be rewritten as:
{\setlength{\abovedisplayskip}{4pt}
\setlength{\belowdisplayskip}{4pt}
\setlength{\abovedisplayshortskip}{2pt}
\setlength{\belowdisplayshortskip}{2pt}
\begin{equation}
    \mathcal{E}(Q) = \mathbbm{1}\!\left[|\mathcal{E}_{\text{fact}}| + |\mathcal{E}_{\text{conflict}}| > 0 \right].
    \label{eq:eq_indicator}
\end{equation}
}

\paragraph{Diagnostic Summary}
Abstract factual judgments are subsequently grounded into concrete natural language claims, ensuring alignment between high-level reasoning and specific factual assertions. The Detect module ultimately produces a structured diagnostic summary:
{\setlength{\abovedisplayskip}{4pt}
\setlength{\belowdisplayskip}{4pt}
\setlength{\abovedisplayshortskip}{2pt}
\setlength{\belowdisplayshortskip}{2pt}
\begin{equation}
    \textsc{MisSum}(Q) = \text{Summarize}\!\left(\mathcal{E}_{\text{fact}} \cup \mathcal{E}_{\text{conflict}}\right),
    \label{eq:mis_summary}
\end{equation}
}
which specifies \textit{whether }errors exist, \textit{where }they occur, and \textit{why }they are problematic, providing an interpretable foundation for strategy formulation.

\subsection{Devise: Multi-Perspective Deliberation}

Even when errors are correctly identified, a single model may still undercorrect by overlooking subtler errors, overcorrect by invalidating valid content, or introduce new inconsistencies during correction. All these failure modes share a common root cause: the model is constrained by its own reasoning frame and cannot adequately challenge its initial assumptions. To address this, we propose a \textbf{multi-perspective strategy debate mechanism} that decomposes correction strategy formulation into three complementary roles.

All three roles condition on the model's internal knowledge and on $\textsc{MisSum}(Q)$, which encodes the factual and conflict errors identified in $Q$.

\paragraph{Generator ($\mathcal{G}$)} 
The generator produces an initial draft strategy $s^{(0)}$ grounded in $\textsc{MisSum}(Q)$, aiming to enumerate all plausible correction pathways:
{\setlength{\abovedisplayskip}{4pt}
\setlength{\belowdisplayskip}{4pt}
\setlength{\abovedisplayshortskip}{2pt}
\setlength{\belowdisplayshortskip}{2pt}
\begin{equation}
    s^{(0)} = \mathcal{G}\!\left(Q, \textsc{MisSum}(Q)\right).
    \label{eq:generator}
\end{equation}
}

\paragraph{Reviewer ($\mathcal{R}$)} 
The reviewer takes an adversarial stance, analogous to a peer reviewer, and identifies weaknesses in $s^{(0)}$ along three dimensions: 
(1) \emph{\textbf{completeness}}, assessing whether all erroneous facts, misleading assumptions, and conflicts in $\mathcal{E}_{\text{fact}} \cup \mathcal{E}_{\text{conflict}}$ are addressed; 
(2) \emph{\textbf{accuracy}}, penalizing overcorrection or unsubstantiated claims; 
(3) \emph{\textbf{reliability}}, evaluating whether the strategy suffices to guide the model toward a correct and on-topic response. It produces a structured critique:
{\setlength{\abovedisplayskip}{4pt}
\setlength{\belowdisplayskip}{4pt}
\setlength{\abovedisplayshortskip}{2pt}
\setlength{\belowdisplayshortskip}{2pt}
\begin{equation}
    r = \mathcal{R}\!\left(s^{(0)},\, \textsc{MisSum}(Q)\right),
    \label{eq:reviewer}
\end{equation}
}
where $r$ encodes any deficiencies in the draft strategy. If no deficiencies are found, $r$ may be empty and $s^{(0)}$ is adopted directly.

\paragraph{Arbiter ($\mathcal{A}$)} 
The arbiter intervenes whenever \( r \neq \varnothing \), i.e., only when the reviewer identifies deficiencies in the draft strategy. Acting without bias toward either party, it impartially evaluates both perspectives and selectively incorporates their valid insights to generate a final strategy that is comprehensive, accurate, and actionable:
{\setlength{\abovedisplayskip}{4pt}
\setlength{\belowdisplayskip}{4pt}
\setlength{\abovedisplayshortskip}{2pt}
\setlength{\belowdisplayshortskip}{2pt}
\begin{equation}
    \pi^*(Q) = \mathcal{A}\!\left(s^{(0)},\, r\right).
    \label{eq:arbiter}
\end{equation}
}
Role separation mitigates self reinforcing bias common in single model self correction.
 The generator 
proposes, the reviewer critiques, and the arbiter resolves disagreements. This produces a multi perspective 
validated strategy \(\pi^{*}(Q)\) before execution.

\subsection{Correct: Correction and Response}

In the final stage, the model executes the validated strategy $\pi^*(Q)$ to generate a response. Even a carefully devised strategy may fail if the model deviates during generation, so we guide the model to follow the strategy's steps sequentially. Each step corresponds to a concrete action in the correction pipeline: (1) Error Identification, explicitly pointing out factual errors in the input; (2) Correction with Justification, providing the correct information along with supporting reasoning; and (3) Reliable Answering, producing a final answer to the original question under the corrected premises.

By following this sequential execution, the model ensures that all identified errors are addressed systematically, that corrections are justified, and that the final response faithfully reflects the validated strategy. 

\subsection{Implementation Strategies}

We implement the DEDUCE framework using two complementary strategies: DEDUCE-Prompting and DEDUCE-Tuning.

\paragraph{DEDUCE-Prompting}
DEDUCE-Prompting decomposes into three modules. The Detect module identifies factual errors and conflicts in the input. The Devise module generates multi-perspective correction strategies using Generator, Reviewer, and Arbiter roles to ensure comprehensive and reliable strategy formulation. The Correct module executes the validated strategy and produces a final answer with justifications.
This strategy is easy to deploy, fully interpretable, and flexible across tasks. It allows direct use of existing language models without additional training. Prompts are provided in Appendix ~\ref{sec:imple_details}

\paragraph{DEDUCE-Tuning}
DEDUCE-Tuning aims to internalize the reasoning patterns of DEDUCE into model parameters, enabling efficient and strategy-faithful correction without multi-step prompting.
\paragraph{Stage 1: Detect Fine-Tuning} 
In the first stage, the model learns to identify factual errors and conflicts. Training data is generated using DEDUCE-Prompting with GPT-4o as the teacher. Each input query is paired with its diagnostic summary and error labels. Only examples where the teacher outputs are fully correct are included. This stage ensures the model develops a reliable foundation for factual error detection.
\vspace{-0.6em}
\paragraph{Stage 2: Devise and Correct Fine-Tuning} 
The second stage integrates the Devise and Correct modules. Training examples are generated using Qwen2.5-14B as the teacher, including validated multi-perspective strategies and the corresponding correct answers. We filter examples based on a strict quality criterion, keeping only those with a clarify score of 5 and verified correctness. By combining Devise and Correct into a single fine-tuning stage, the model internalizes multi-perspective reasoning and executes corrections in one step, mitigating the risk of self-reinforcing biases.
Further implementation details are provided in Appendix ~\ref{sec:deduce}.

\section{Benchmark}
\paragraph{Terminology} We extend the term \textit{clarification} beyond resolving ambiguity 
to include the process by which the model identifies factual errors in user inputs and provides corrected information.

\subsection{Benchmark Metrics} \label{sec:bench-metrics}
To evaluate model behavior on queries containing factual errors, we introduce three complementary metrics: \textbf{Misleading Rate}, \textbf{Correction Rate}, and \textbf{Clarification Score}. Together, these metrics assess whether a model (i) is misled by erroneous inputs, (ii) corrects the underlying inaccuracies, and (iii) provides reliable answer.

Existing evaluation practices are insufficient for this setting. Traditional metrics such as BLEU~\cite{papineni-etal-2002-bleu} and ROUGE~\cite{lin2004rouge} measure surface level textual similarity and cannot evaluate factual correctness. Accuracy only captures whether the final answer is correct, but model responses often exhibit varying degrees of correctness rather than being entirely right or wrong. Our metrics address these gaps by measuring susceptibility to misleading inputs, ability to recover correct facts, and partial correctness.

We define the set of erroneous queries as:
{\setlength{\abovedisplayskip}{4pt}
\setlength{\belowdisplayskip}{4pt}
\setlength{\abovedisplayshortskip}{2pt}
\setlength{\belowdisplayshortskip}{2pt}
\begin{equation}
\textstyle
\mathcal{Q}_{\text{err}}
= \{\, Q \mid \mathcal{E}(Q) = 1\}.
\label{eq:ierr}
\end{equation}
}

\paragraph{Misleading Rate (MR)}
This metric measures the model's susceptibility to input-induced hallucinations. It is defined as the proportion of erroneous queries where the model fails to recognize the error and instead adopts the user's false premise.

{\setlength{\abovedisplayskip}{4pt}
\setlength{\belowdisplayskip}{4pt}
\setlength{\abovedisplayshortskip}{2pt}
\setlength{\belowdisplayshortskip}{2pt}
\begin{equation}
\textstyle
\mathrm{MR}
= \frac{
\sum_{Q \in \mathcal{Q}_{\text{err}}} L(Q)
}{
\lvert \mathcal{Q}_{\text{err}} \rvert
},
\label{eq:mr}
\end{equation}
}
where $L(Q)=1$ if the response accepts the false premise, $0$ otherwise.

\paragraph{Correction Rate (CR)}
This metric measures the model's ability to detect errors. It calculates the proportion of erroneous inputs where the model identifies most false or contradictory claims. If the model recognizes most errors, it is considered to have error detection capability.
{\setlength{\abovedisplayskip}{4pt}
\setlength{\belowdisplayskip}{4pt}
\setlength{\abovedisplayshortskip}{2pt}
\setlength{\belowdisplayshortskip}{2pt}
\begin{equation}
\textstyle
\mathrm{CR}
=
\frac{
\sum_{Q \in \mathcal{Q}_{\text{err}}} R(Q)
}{
\lvert \mathcal{Q}_{\text{err}} \rvert
},
\label{eq:cr}
\end{equation}
}
where $R(Q)=1$ if the response corrects the errors, and $0$ otherwise.

\paragraph{Clarification Score(CS)}
The Clarification Score measures how effectively a model identifies and addresses factual errors in a query. The specific definition is shown in Table \ref{tab:clarification-score}.

\begin{table}[h]
\centering
\small
\renewcommand{\arraystretch}{1.2} 
\begin{tabular}{c p{0.75\linewidth}} 
\hline
\textbf{Score} & \textbf{Description} \\
\hline
\textbf{1} & \textbf{Misled}: The model accepts the false claims and reinforces the error. \\
\textbf{2} & \textbf{Avoidance}: The model ignores the error and answers directly, resulting in an incorrect response.\\
\textbf{3} & \textbf{Contradictory}: The model identifies some false claims but gives contradictory responses.\\
\textbf{4} & \textbf{Partial Correction}: The model identifies most false claims but does not adequately answer the question under the correct premise.\\
\textbf{5} & \textbf{Full Clarification}: The model explicitly identifies the error, refutes it with correct facts, and provides the accurate answer. \\
\hline
\end{tabular}
\caption{Clarification Score Rubric for evaluating model responses to erroneous queries.}
\label{tab:clarification-score}
\vspace{-15pt}
\end{table}

We define $L(Q)=1$ when CS $\in \{1, 2\}$, indicating the model is completely misled, and $R(Q)=1$ when CS $\in \{4, 5\}$, indicating adequate error correction capability.

\subsection{FPQA Dataset}\label{sec:data-construction}

We construct \textsc{MisFactQA} for the FPQA task by combining three sources with verified ground truth: a subset of Prize~\cite{yuan-etal-2024-whispers}, EchoMist~\cite{guo2025protect}, and publicly available question answering datasets. We create query-response pairs that require models to both identify false information in the input and generate the corresponding correction and answer.

Existing work primarily focuses on single error in the input and evaluates models based only on accuracy, which is insufficient to measure robustness under factual perturbed inputs. To address this limitation, we introduce three types of errors that test factual accuracy, consistency, and comprehensiveness. Queries with a single false premise can mislead models to reason from an incorrect starting point. Queries with internally contradictory descriptions force the model to choose between conflicting claims, making it prone to inconsistent reasoning. Queries containing multiple errors, such as several false premises or a false premise combined with a contradiction, may cause the model to detect one error while overlooking others, further challenging its reasoning process. Further construction details are provided in the Appendix ~\ref{sec:fpqa}.

\begin{table*}[ht!]
\centering
\small
\setlength{\tabcolsep}{5.5pt}
\begin{tabular}{llcccccccc}
\toprule
\multirow{2}{*}{\textbf{Model}} & \multirow{2}{*}{\textbf{Method}} &
\multicolumn{4}{c}{\textbf{FalseQA}} &
\multicolumn{4}{c}{\textbf{MisFactQA}} \\
\cmidrule(lr){3-6} \cmidrule(lr){7-10}
& & \textbf{Acc $\uparrow$} & \textbf{CS $\uparrow$} & \textbf{MR $\downarrow$} & \textbf{CR $\uparrow$}
  & \textbf{Acc $\uparrow$} & \textbf{CS $\uparrow$} & \textbf{MR $\downarrow$} & \textbf{CR $\uparrow$} \\
\midrule

\multirow{7}{*}{Qwen2.5-7B-Instruct}
& Original  & 51.64 & 3.44 & 44.15 & 54.13 & 39.04 & 3.22 & 51.52 & 43.49 \\
& ICL       & 53.25 & 3.58 & 39.36 & 57.45 & 49.36 & 3.56 & 37.48 & 52.83 \\
& CoT       & 52.61 & 3.56 & 40.48 & 55.84 & 45.35 & 3.36 & 46.10 & 50.37 \\
& SFT       & 65.74 & 4.18 & 20.47 & 74.60 & 61.19 & 3.94 & 23.47 & 65.16 \\
& IAQ-FA    & 60.81 & 3.95 & 31.29 & 67.55 & 44.69 & 3.40 & 44.69 & 50.37 \\
\rowcolor{oursbg}
& DEDUCE-P  & \underline{70.59} & \underline{4.29} & \textbf{19.12} & \underline{78.53}
            & \underline{61.24} & \textbf{4.08} & \textbf{18.23} & \underline{70.62} \\
\rowcolor{oursbg}
& DEDUCE-T  & \textbf{77.84} & \textbf{4.32} & \underline{19.72} & \textbf{78.56}
            & \textbf{64.65} & \textbf{4.08} & \underline{21.48} & \textbf{75.20} \\
\midrule

\multirow{7}{*}{LlaMA-3.1-8B-Instruct}
& Original  & 43.75 & 3.10 & 54.84 & 42.66 & 30.16 & 3.01 & 57.49 & 36.23 \\
& ICL       & 57.24 & 3.74 & 33.23 & 63.46 & 63.98 & 4.15 & 18.20 & 73.55 \\
& CoT       & 31.12 & 2.62 & 66.83 & 27.59 & 46.53 & 3.34 & 45.94 & 48.51 \\
& SFT       & 48.39 & 3.37 & 44.81 & 52.09 & 44.59 & 3.98 & 15.41 & 74.31 \\
& IAQ-FA    & 59.29 & 3.78 & 36.23 & 62.62 & 33.27 & 3.36 & 37.28 & 50.10 \\
\rowcolor{oursbg}
& DEDUCE-P  & \underline{68.97} & \underline{4.10} & \underline{24.06} & \underline{71.32}
            & \underline{70.01} & \underline{4.35} & \underline{10.87} & \underline{79.35} \\
\rowcolor{oursbg}
& DEDUCE-T  & \textbf{74.09} & \textbf{4.18} & \textbf{21.00} & \textbf{76.26}
            & \textbf{70.60} & \textbf{4.36} & \textbf{10.64} & \textbf{81.82} \\
\midrule

\multirow{7}{*}{Gemma3-12B-Instruct}
& Original  & 37.02 & 2.78 & 63.77 & 34.81 & 24.20 & 2.68 & 71.91 & 26.15 \\
& ICL       & 42.91 & 3.02 & 56.37 & 42.09 & 36.24 & 3.11 & 52.29 & 41.97 \\
& CoT       & 41.28 & 2.89 & 60.03 & 36.68 & 35.27 & 3.03 & 56.73 & 37.82 \\
& SFT       & 53.60 & 3.54 & 41.27 & 57.14 & 37.52 & 3.36 & 39.43 & 53.90 \\
& IAQ-FA    & 51.95 & 3.47 & 46.25 & 53.04 & 26.06 & 2.73 & 66.79 & 30.50 \\
\rowcolor{oursbg}
& DEDUCE-P  & \underline{73.34} & \underline{4.20} & \underline{22.70} & \underline{75.77}
            & \underline{57.75} & \underline{3.93} & \textbf{20.07} & \underline{64.43} \\
\rowcolor{oursbg}
& DEDUCE-T  & \textbf{79.59} & \textbf{4.26} & \textbf{20.41} & \textbf{77.71}
            & \textbf{64.96} & \textbf{3.94} & \underline{26.64} & \textbf{67.15} \\
\bottomrule
\end{tabular}

\caption{
Main results on FalseQA and MisFactQA.
Clarification Score is evaluated on a 1--5 scale. $\uparrow$ indicates higher is better, and $\downarrow$ indicates lower is better. Best results are shown in \textbf{bold}, and second-best results are \underline{underlined}.}
\label{tab:main-results}
\vspace{-10pt}
\end{table*}

\section{Experiments}
In this section, we conduct comprehensive experiments to verify the effectiveness of DEDUCE. Specifically, we aim to address the following research questions: \textbf{RQ1 (Effectiveness):} How well does our framework perform compared with baseline methods? \textbf{RQ2 (Error Analysis):} How do various types of factual errors influence model behavior? \textbf{RQ3 (Component Contribution):} How does each component of the proposed method contribute to overall performance? \textbf{RQ4 (Robustness and Efficiency):} Does the vulnerability to fact perturbed inputs persist in stronger LLMs, and what is the inference cost of applying DEDUCE?

\subsection{Experimental Setup}

\paragraph{Datasets and Evaluation Metrics}
We evaluate DEDUCE on three benchmark datasets: \textbf{TruthfulQA}~\citep{lin-etal-2022-truthfulqa}, which contains misleading questions; \textbf{FalseQA}~\citep{hu-etal-2023-wont}, a benchmark for detecting false premises; and \textbf{MisFactQA}, the dataset we construct in Section~\ref{sec:data-construction} to assess model performance under diverse factual errors.

Following previous studies, we evaluate all models using accuracy (Acc), where a model's response is considered correct only if it contains the ground truth answer. We additionally report three metrics: Misleading Rate (MR), measuring how often models are completely misled; Correction Rate (CR), reflecting a reasonably good ability to correct errors; and Clarification Score (CS), a 1-5 scale rating ranging from fully misled to complete correction with accurate responses.
\paragraph{Baseline Methods}
We compare DEDUCE with five representative categories of baselines:
\textbf{Default Model:} The original LLM with a standard system prompt.
\textbf{ICL}~\citep{brown2020language}\textbf{:} In context learning with demonstration examples.
\textbf{CoT}~\citep{wei2022chain}\textbf{:} Chain-of-thought prompting for step-by-step reasoning.
\textbf{LoRA Fine Tuning}~\citep{hu2022lora}\textbf{:} Parameter efficient supervised fine tuning.
\textbf{IAQ FA}~\citep{wang-blanco-2025-identifying}: An interpretable false assumption detection method.

\paragraph{Models and Implementation Details}
We employ models from the Qwen~\citep{qwen2025qwen25technicalreport}, Llama~\citep{grattafiori2024llama}, and Gemma~\citep{team2024gemma} families. Experiments are conducted using Ollama\footnote{\url{https://ollama.com/}} with temperature = 0.0. Following the prior work~\cite{li-etal-2025-dont}, we use o3-mini-medium as an automated judge for evaluation, given its strong performance on reasoning and knowledge-intensive tasks in JudgeBench~\citep{tan2024judgebench}. Comprehensive details regarding datasets, baselines, and implementation settings are provided in Appendix ~\ref{sec:appendix}. 
We further assessed the reliability of the automated scoring process through human evaluation. The LLM judge shows near-perfect agreement with human judgments on accuracy, with Cohen's $\kappa=0.936$, and strong alignment on the clarification score, with Pearson's correlation coefficient $r=0.916$. These results support the validity of our evaluation metrics. Additional details are provided in Appendix~\ref{sec:evaluation agreement}.

\subsection{Overall Performance}

To address RQ1, we evaluate the performance of our proposed method against the baselines. The main results on FalseQA and MisFactQA are presented in Table~\ref{tab:main-results}. DEDUCE consistently outperforms all baseline methods across all models and datasets. On the Gemma3-12B model, our approach improves upon the best baseline by a substantial margin of 25.99\% in accuracy on FalseQA. Furthermore, DEDUCE achieves the highest correction rate while maintaining the lowest misleading rate. Similar improvements are observed on MisFactQA, demonstrating that our method effectively enhances factual accuracy and the model's ability to identify and correct errors.

An important finding is that, on the TruthfulQA dataset, CoT underperforms the original baseline in several cases, as shown in Table~\ref{tab:truthfulqa_main_ablation}. This counterintuitive result validates our hypothesis: \textit{when inputs contain errors and the model fails to recognize them, reasoning based on the flawed information amplifies the errors, leading to incorrect responses}. This further underscores the necessity of detecting fact perturbations in inputs.

\begin{table}[h]
\centering
\small
\resizebox{0.48\textwidth}{!}{
\setlength{\tabcolsep}{6pt}
\begin{tabular}{lccc}
\toprule
\textbf{Method} & \textbf{Qwen2.5-7B} & \textbf{LlaMA-3.1-8B} & \textbf{Gemma3-12B}\\
\midrule
Original        & 67.11 & 48.67 & 63.74 \\
ICL             & \underline{69.44} & \underline{64.77}& 64.91 \\
COT             & 61.84 & 47.98 & 61.11 \\
SFT             & 67.69 & 52.49 & \underline{66.08}\\
IAQ-FA          & 63.88 & 60.41& 60.68 \\
\rowcolor{lightpurple}
DEDUCE-P            & \textbf{70.47} & \textbf{67.06} & \textbf{74.56} \\
\bottomrule
\end{tabular}}
\caption{TruthfulQA (Multi-Choice) accuracy (\%) across different Instruct models and methods. Best is \textbf{bold}, second-best is \underline{underlined}. }
\label{tab:truthfulqa_main_ablation}
\vspace{-15pt}
\end{table}


Experiments across different model families and scales show that DEDUCE brings significant and consistent performance gains, confirming its effectiveness and strong scalability.

\begin{table*}[h]
\centering
\resizebox{0.8\textwidth}{!}{
\setlength{\tabcolsep}{5.5pt}
\begin{tabular}{lccccccccc}
\toprule
\multirow{2}{*}{Method}& \multicolumn{1}{c}{TruthfulQA (MC)}
& \multicolumn{4}{c}{MisFactQA}
& \multicolumn{4}{c}{FalseQA} \\
\cmidrule(lr){2-2}
\cmidrule(lr){3-6}
\cmidrule(lr){7-10}
& Acc$\uparrow$
& Acc$\uparrow$ & MR$\downarrow$ & CR$\uparrow$ & CS$\uparrow$
& Acc$\uparrow$ & MR$\downarrow$ & CR$\uparrow$ & CS$\uparrow$ \\
\midrule
Original
& 48.67
& 30.16 & 57.49 & 36.23 & 3.01
& 43.75 & 54.84 & 42.66 & 3.10 \\

\rowcolor{gray!10}
Ours (base)
& \textbf{67.06}& \textbf{70.01}& \textbf{10.87}& \textbf{79.35}& \textbf{4.35}& \textbf{68.97}& \textbf{24.06}& \textbf{71.32}& \textbf{4.10}\\

w/o $\textsc{Strategy}(Q)$
& \underline{63.77}
& 44.75 & 17.21 & 70.28 & 3.93
& 60.73 & 39.65 & 58.09 & 3.61 \\

w/o $\textsc{MisSum}(Q)$
& 62.76
& \underline{65.57} & \underline{12.02}& \underline{77.60} & \underline{4.28}
& \underline{67.87} & \underline{26.18} & \underline{69.73} & \underline{4.03} \\
\bottomrule
\end{tabular}}
\vspace{-2pt}
\caption{Ablation Results (Absolute Scores) on LlaMA-3.1-8B-Instruct. Clarification Score is evaluated on a 1–5 scale. ↑ indicates higher is better, and ↓ indicates lower is better. Best is \textbf{bold}, second-best is \underline{underlined}.}
\label{tab:ablation-module}
\vspace{-4pt}
\end{table*}

\begin{table*}[t]
\centering
\small
\setlength{\tabcolsep}{5pt}
\begin{tabular}{llrrrr}
\toprule
\multirow{2}{*}{Models} & \multirow{2}{*}{Method} 
& \multicolumn{2}{c}{FalseQA} 
& \multicolumn{2}{c}{MisFactQA} \\
\cmidrule(lr){3-4} \cmidrule(lr){5-6}
& & Acc. (\%) & Avg. tok. & Acc. (\%) & Avg. tok. \\
\midrule

\multirow{5}{*}{GPT-4o-mini}
& Original & 65.2 & 14.6  & 65.1 & 27.9  \\
& ICL      & 74.6 & 17.3  & 70.1 & 34.2  \\
& CoT      & 62.8 & 125.2 & 62.9 & 141.4 \\
& IAQ-FA   & 82.8 & 115.5 & 65.3 & 156.9 \\
& DEDUCE-T   & 83.9 & 74.1  & 82.9 & 90.0  \\

\midrule

\multirow{5}{*}{DeepSeek-V3}
& Original & 69.5 & 10.4  & 69.9 & 24.7  \\
& ICL      & 79.1 & 124.1 & 83.8 & 154.2 \\
& CoT      & 71.4 & 172.8 & 70.2 & 216.4 \\
& IAQ-FA   & 79.0 & 110.1 & 78.0 & 151.5 \\
& DEDUCE-T   & 87.6 & 73.5  & 84.6 & 100.8 \\

\bottomrule
\end{tabular}
\caption{Performance comparison and Efficiency analysis on stronger LLMs. }
\label{tab:qa_results}
\end{table*}
\vspace{-4pt}

\vspace{-5pt}
\subsection{Error Analysis}
To address RQ2, we analyze Qwen2.5-7B and LlaMA-3.1-8B across four query types: (1) Correct QA with true premises; (2) Contradictory Premise with internally conflicting descriptions; (3) False Premise built on incorrect assumptions; and (4) Complex Error combining multiple error types. To disentangle failures caused by knowledge gaps from those triggered by misleading inputs, these questions are all formulated from the same underlying background information. 

As shown in Figure~\ref{fig:misleading performance}, both models perform well on correct questions but degrade substantially when errors are introduced. Qwen2.5-7B drops from 75\% to 45\% (Contradictory), 15\% (False Premise), and 25\% (Complex Error); LlaMA-3.1-8B-Instruct shows similar patterns, declining from 62\% to 32\%, 16\%, and 19\%, respectively. Notably, error detectability varies by type: contradictory premises present overt conflicts that models detect more easily, whereas false premises appear superficially plausible, making them harder to identify. Additional scaling results across model families, sizes, and prompting styles are provided in Appendix ~\ref{error analysis}.

\subsection{Component Contribution Analysis}

We conduct ablation experiments to address RQ3 and evaluate the contribution of the core components in our framework: (1) the Detect module ($\textsc{MisSum}(Q)$), and (2) the Devise module ($\textsc{Strategy}(Q)$). As shown in Table~\ref{tab:ablation-module}, experiments on LlaMA-3.1-8B-Instruct across TruthfulQA, MisFactQA, and FalseQA demonstrate that each module contributes positively to overall performance, and removing any component leads to degradation. The Strategy module has the largest impact. This suggests that merely detecting input errors is insufficient; tailored correction strategies are crucial for substantial performance gains.

We further investigate how the Generator, Reviewer, and Arbiter collaborate to enhance response quality, as well as how the depth of their interaction affects the effectiveness of the strategy. This multi-perspective reasoning mechanism enables each role to identify potential flaws in the strategy, refine the reasoning process, and ultimately develop a more effective error-correction strategy.
 Experimental results show that as the number of discussion rounds increases, most models become more robust to misleading inputs. When the initial strategy is already of high quality, a single round of discussion or direct generation is sufficient; for more challenging queries, deeper deliberation can further optimize strategy quality. Detailed analysis in Appendix ~\ref{rounds}.

\subsection{Generalization to Stronger LLMs and Efficiency Analysis}

To further examine whether fact perturbations remain a critical challenge for stronger LLMs and evaluate the inference cost introduced by DEDUCE, we conduct additional experiments on GPT-4o-mini and DeepSeek-V3. The results in Table~\ref{tab:qa_results} demonstrate that stronger models remain vulnerable to misleading factual inputs. On MisFactQA, the original GPT-4o-mini and DeepSeek-V3 achieve only 65.1\% and 69.9\% accuracy, respectively. With DEDUCE, the accuracy improves to 82.9\% and 84.6\%, respectively. Similar improvements are observed on FalseQA, demonstrating that stronger model capabilities do not fully eliminate the impact of erroneous user premises. We further analyze inference cost by measuring the average number of generated tokens. While DEDUCE introduces additional token usage compared with simple baselines (Original and ICL), it achieves substantially higher accuracy. Compared with robustness-oriented methods (CoT and IAQ-FA), DEDUCE achieves a better accuracy-efficiency trade-off, providing stronger robustness with lower token consumption. These results demonstrate that DEDUCE achieves robustness improvements through its structured error detection and correction process under a reasonable inference cost.

\vspace{-5pt}
\subsection{Case Study}
\vspace{-3pt}
From case studies on TruthfulQA, FalseQA, and MisFactQA (see Appendix~\ref{sec:case-study}), we observe that LLM failures under misleading queries largely stem from the uncritical acceptance of erroneous information in the input. Baseline models tend to follow these errors and produce incorrect answers. In contrast, DEDUCE prevents error propagation through fine-grained decomposition of inputs into atomic claims and explicit identification of invalid assumptions, enabling the model to clarify misconceptions and answer correctly. For example, as shown in the MisFactQA case (Table~\ref{tab:case-Misfact}), when a query contains multiple false statements regarding Nobel Prize categories and years, our framework successfully identifies all major errors and generates a factually accurate, corrected response. This ability to recognize errors enables more reliable and factually grounded responses.

\vspace{-5pt}
\section{Related Work}
\vspace{-5pt}
\paragraph{Input Induced Hallucinations}
Erroneous user inputs, ranging from explicit factual inaccuracies to complex misconceptions, trigger input induced hallucinations where models generate plausible but incorrect responses by conforming to flawed queries. Recent studies attribute this vulnerability to two key phenomena: sycophancy, where models align with user views to maximize perceived helpfulness~\citep{pmlr-v235-chen24u, malmqvist2025sycophancy}, and knowledge conflict, where misleading external context overrides internal parametric knowledge~\cite{xu2024knowledge, su2024conflictbank, zhang2025faithfulrag, wang-etal-2025-astute, jin2024cutting, tan2024blinded}. While prior work has developed methods for detecting false premises to ensure robust answering~\citep{wang-blanco-2025-identifying, yuan-etal-2024-whispers}, these approaches often neglect the critical need to explicitly address and correct the user's underlying misconception. Our work directly addresses this gap by not only identifying input errors but also actively clarifying them, thereby resolving knowledge conflicts and preventing the reinforcement of user misunderstandings.
\vspace{-5pt}
\paragraph{Hallucination Mitigation Methods}
Existing hallucination mitigation research primarily targets errors originating from the model's generation process. Inference-time techniques, notably Retrieval-Augmented Generation, enhance factuality by grounding responses in external knowledge sources~\citep{peng2023check, izacard2023atlas, lyu2024retrieve}, while specialized decoding strategies reduce fabrication by constraining uncertain tokens~\citep{lee2022factuality, dhuliawala2024chain}. Post-hoc methods employ fact-checking or self-correction pipelines to identify and edit errors in generated text~\citep{zhang2024self}. However, these methods typically assume user input is correct, thereby overlooking the risk of input-induced hallucinations. They focus on verifying output rather than challenging input premises.


\vspace{-5pt}
\paragraph{Self Correction Methods}  
Self-correction mechanisms, such as Self Refine~\cite{madaan2023self} and CRITIC~\citep{gou2023critic}, enable large language models to iteratively refine their outputs based on feedback. However, most existing self-correction approaches primarily focus on verifying the final answer. In contrast, our approach applies a critique mechanism specifically to the input analysis and response strategy phases, validating premises before generation.

\vspace{-5pt}
\section{Conclusion}
\vspace{-5pt}
This paper highlights the critical yet overlooked problem of fact perturbations in user inputs that systematically induce hallucinations in LLMs. To address this, we propose DEDUCE, a framework that enhances model robustness by detecting, devising, and correcting such input errors. We contribute the MisFactQA dataset and fine-grained metrics to evaluate model behavior under unreliable queries. Comprehensive experiments validate DEDUCE's effectiveness across multiple benchmarks and model families. Our work shifts the focus towards building more vigilant and reliable LLMs that can actively validate and correct user premises. 

\section*{Limitations}

\begin{enumerate}
    \item \textbf{Model scale constraints.} While our experimental evaluation centers on mid-scale models including Qwen2.5-7B, LlaMA-3.1-8B, and Gemma3-12B, we plan to extend our analysis to a broader range of models to validate the generalizability of our approach and identify potential model-specific optimizations.
    
    \item \textbf{Task scope focused on factual errors.} Our work specifically targets queries with factual inaccuracies or contradictions (fact perturbations). Other forms of unreliable inputs, such as ambiguous queries or adversarially crafted prompts, are not explored in this work.
\end{enumerate}

\section*{Ethics Statement}
This research proposes the DEDUCE framework to enhance large language models' ability to identify and correct factual errors in user inputs, aiming to improve the reliability of AI systems in high-stakes domains such as healthcare and law, thereby providing positive societal value. The adversarial argument generation involved in the method is solely used for model robustness verification and does not produce misleading content. All datasets used are publicly available resources, contain no personal privacy or sensitive information, and comply with academic standards. In accordance with conference policy, we acknowledge the use of AI tools for text polishing, and the authors take full responsibility for the content, ensuring the originality and validity of the research.

\bibliography{reference}

\clearpage
\appendix

\section{Experimental Details}
\label{sec:appendix}

\subsection{Dataset details}
\paragraph{TruthfulQA}~\cite{lin-etal-2022-truthfulqa}
TruthfulQA-MC is a benchmark designed to evaluate whether language models produce truthful answers in the presence of common human misconceptions.
It comprises 684 questions spanning 38 categories, including health, law, finance, and politics.
We use the simplified multiple-choice version of TruthfulQA provided by prior work, in which the original dataset’s variable-length multiple-choice questions are standardized by removing questions with fewer than four options and randomly sampling four choices for the remaining questions.
\paragraph{FalseQA}~\cite{hu-etal-2023-wont} We use the FalseQA dataset to evaluate models' ability to handle false-premise questions. The dataset contains 2,365 FPQ-TPQ pairs, where each false-premise question (FPQ) is annotated with an explanation of the incorrect assumption and a revised true-premise question (TPQ). It covers 8 error types (Property, Action, Scope, Entity, Event, Logic, Causality, Index) and 6 question forms (Descriptive, Factual, Enumerative, Selective, Hypothetical, Affirmative). 

\paragraph{MisFactQA}\label{misfactqa}
We primarily utilize a subset of Prize~\cite{yuan-etal-2024-whispers}, EchoMist~\cite{guo2025protect}, and a publicly available Hugging Face dataset, \texttt{hallucinations-dpo}. We further apply filtering, reconstruction, and data augmentation to these datasets. 
Using the data construction method described in Appendix ~\ref{sec:fpqa}, we build a dataset that spans a spectrum of error complexity, including queries with a single erroneous claim as well as queries containing multiple co-occurring factual errors, such as false premises, internal contradictions, and their combinations. 

The Prize dataset centers on Nobel Prize related topics and is used to construct questions grounded in the same knowledge background but exhibiting different types of errors. EchoMist and the web data consist of real world conversations and simulated data covering topics such as health, fake news, food, technology, conspiracy, folklore, medicine, science, politics, lifestyle, and history.  Finally, we obtain 900 pairs of questions that share the same underlying information but exhibit different error types. In addition, we construct 1,140 questions with fact perturbations, where multiple types of errors are simultaneously introduced. 

Overall, the resulting dataset reflects complex scenarios in which user inputs contain factual inaccuracies, factual contradictions, and multiple errors occurring simultaneously. \textit{The dataset is designed to evaluate the robustness and reliability of models in the presence of fact perturbations}, specifically whether a model can detect potential errors in the input, correct erroneous information, and provide reliable answers without being misled by such errors. Detailed examples are shown in Table~\ref{tab:misfact-examples}.

Detailed statistics of the datasets used in our experiments are shown in Table~\ref{tab:dataset_stats}.

\begin{table}[h]
\centering
\setlength{\tabcolsep}{10pt}
\begin{tabular}{lr}
\toprule
\textbf{Dataset} & \textbf{Nums} \\
\midrule
TruthfulQA(MC)     & 684 \\
FalseQA  & 3274 \\
MisFactQA   & 1140 \\
\bottomrule
\end{tabular}
\caption{Statistics of datasets for experiments.}
\label{tab:dataset_stats}
\vspace{-10pt}
\end{table}

\subsection{Implementation Details} \label{sec:imple_details}
\begin{figure*}[t]
  \centering
  \includegraphics[width=1\textwidth]{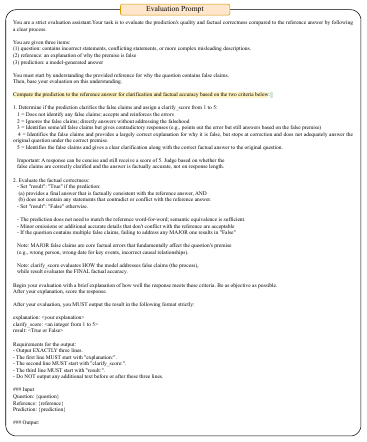}
  \caption{Evaluation prompt used for assessing model responses to FPQA.}
  \label{fig:eval-prompt}
\end{figure*}

We employ LlaMA-3.1-8B-Instruct, Qwen2.5-7B-Instruct and Gemma3-12B-Instruct as backbone LLMs and conduct experiments using Ollama\footnote{Ollama: \url{https://ollama.com/}.}
. For all models we set temperature=0.0. Based on JudgeBench~\cite{tan2024judgebench}, we evaluate model responses using \textbf{o3-mini-medium} as an automated judge, given its strong performance on reasoning and knowledge-intensive evaluation tasks. The reported results are based on a single run.
\paragraph{Evaluation Prompts} 
Detailed evaluation prompt are provided in Figure~\ref{fig:eval-prompt}.
\paragraph{Prompts for the DEDUCE framework} 

The DEDUCE framework employs a structured, multi-stage process to detect, analyze, and correct factual errors in user queries. This process is orchestrated through a series of carefully designed prompts, each tailored for a specific task in the pipeline. 

\begin{figure*}[h]
  \centering
  \includegraphics[width=1\textwidth]{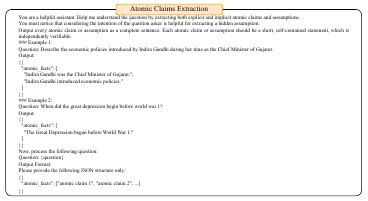}
  \caption{Prompt used for Atomic Claims Extraction.}
  \label{fig:AC-Extraction}
\end{figure*}

\begin{figure*}[h]
  \centering
  \includegraphics[width=1\textwidth]{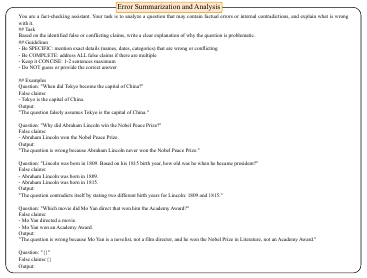}
  \caption{Prompt used for Error Summarization.}
  \label{fig:Err-Summary}
\end{figure*}

\begin{figure*}[h]
  \centering
  \includegraphics[width=1\textwidth]{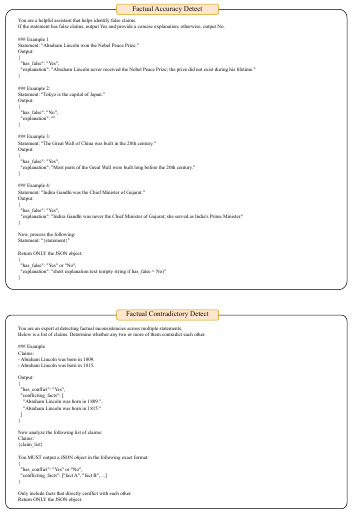}
  \caption{Prompt used for Factual Detection.}
  \label{fig:AC-Detect}
\end{figure*}

\begin{figure*}[h]
  \centering
  \includegraphics[width=1\textwidth]{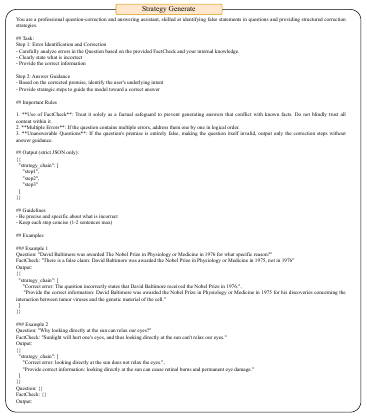}
  \caption{Prompt used for Strategy Generation.}
  \label{fig:strategy-generate}
\end{figure*}

\begin{figure*}[h]
  \centering
  \includegraphics[width=1\textwidth]{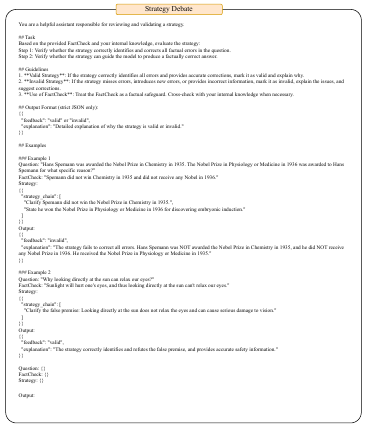}
  \caption{Prompt used for Strategy Debate.}
  \label{fig:strategy-debate}
\end{figure*}

\begin{figure*}[h]
  \centering
  \includegraphics[width=1\textwidth]{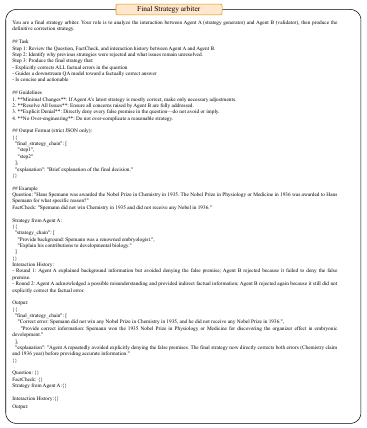}
  \caption{Prompt used for Final Strategy Arbiter.}
  \label{fig:strategy-arbiter}
\end{figure*}

\begin{figure*}[h]
  \centering
  \includegraphics[width=1\textwidth]{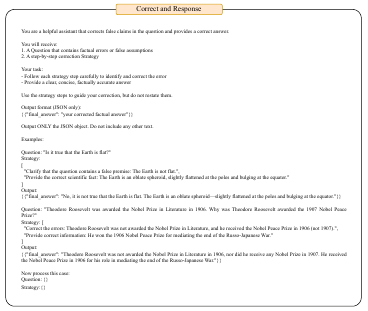}
  \caption{Prompt used for Correct and Response.}
  \label{fig:answer}
\end{figure*}

The process begins by decomposing the user's query into minimal, verifiable statements using the \textbf{Atomic Claims Extraction} prompt (Figure~\ref{fig:AC-Extraction}). Each extracted claim is then rigorously evaluated for factual accuracy and internal consistency by the \textbf{Factual Verification} prompt (Figure~\ref{fig:AC-Detect}). If any inaccuracies or contradictions are found, the \textbf{Error Summarization and Analysis} prompt (Figure~\ref{fig:Err-Summary}) is used to generate a concise diagnostic summary of the issues.
Following the error analysis, the framework enters a collaborative strategy formulation phase. An initial correction plan is proposed via the \textbf{Strategy Generation} prompt (Figure~\ref{fig:strategy-generate}). This proposal is then critically examined during the \textbf{Strategy Debate} stage (Figure~\ref{fig:strategy-debate}). When the maximum number of interaction rounds is reached without consensus, a Final strategy is established by the \textbf{Final Strategy Arbiter}. (Figure~\ref{fig:strategy-arbiter}). Finally, the validated plan is executed using the \textbf{Correct and Response} prompt (Figure~\ref{fig:answer}), which guides the model to generate a factually accurate and reliable answer.

\subsection{Baselines details}

\paragraph{SFT}  We employ LoRA for efficient fine-tuning. The detailed setting of hyperparameters is shown in Table~\ref{tab:train_config}. The parameter settings for the SFT baseline and Deduce-Tuning are identical.

\begin{table}[h]
\centering
\begin{tabular}{@{}ll@{}}
\toprule
\textbf{Configuration} & \textbf{Value} \\
\midrule
Number of epochs & 2 \\
Devices & 3~$\times$~A40 GPUs \\
Total batch size & 32 \\
Learning rate & $2 \times 10^{-5}$ \\
Warmup ratio & 0.1 \\
\bottomrule
\end{tabular}
\caption{Training configuration.}
\label{tab:train_config}
\end{table}

Detailed statistics of the datasets used for training and testing are presented in Table~\ref{tab:dataset_info}.
\begin{table}[ht]
\centering
\begin{tabular}{lcc}
\toprule
\textbf{Dataset} & \textbf{Train} & \textbf{Test} \\
\midrule
FalseQA(FPQ)& 1187& 687\\
FalseQA(TPQ)& 1187& 213\\
TruthfulQA &   0& 684\\
MisFactQA& 540& 600\\
\bottomrule
\end{tabular}
\caption{Dataset information with train and test set sizes.}
\label{tab:dataset_info}
\end{table}

\paragraph{IAQ-FA} 
{IAQ-FA}~\cite{wang-blanco-2025-identifying}: An interpretable 
false assumption detection method. We use GPT-4o for atomic assumption 
extraction and verification without RAG for fair comparison, and extend 
it to generate answers based on the detection results.

\section{Method Details}

\subsection{DEDUCE-T Implementation Strategies} \label{sec:deduce}

\paragraph{Data Formalization} 
We construct two stage specific training datasets to match the DEDUCE-Tuning process:
\begin{equation}
\small
\mathcal{D}_\text{Detect} = \left\{ \big(Q^{(i)}, \text{MisSum}(Q^{(i)}), y_\text{error}^{(i)}\big) \right\}_{i=1}^{N_\text{Detect}}
\label{eq:dataset-detect}
\end{equation}

\begin{equation}
\small
\mathcal{D}_\text{Devise+Correct} = \left\{ \big(Q^{(i)}, y_\text{Correct}^{(i)}\big) \right\}_{i=1}^{N_\text{DevCorrect}}
\label{eq:dataset-devcorrect}
\end{equation}
For the Detect stage, each example consists of the input query \(Q\), the diagnostic summary \(\text{MisSum}(Q)\), and a binary error label \(y_\text{error}\). For the Devise and Correct stage, each example consists of the input query \(Q\) and the final corrected answer \(y_\text{Correct}\).  

This design provides the model with the necessary context at each stage, enabling it to learn reliable error detection in the first stage and internal strategy-guided correction in the second stage.

\paragraph{Training Objective}

\begin{equation}
\small
\mathcal{L}_\text{Detect}(\theta) = - \sum_{\mathcal{D}_\text{Detect}} 
\log P_\theta(y_\text{error}, \text{MisSum(Q)} \mid Q)
\label{eq:loss-detect}
\end{equation}
In the detect stage, the model learns to produce a diagnostic summary and predict whether each atomic fact contains an error.

\begin{equation}
\small
\mathcal{L}_\text{Devise+Correct}(\theta) = - \sum_{\mathcal{D}_\text{Devise+Correct}}
\log P_\theta(y_\text{Correct} \mid Q)
\label{eq:loss-devcorrect}
\end{equation}
In the second stage, the model performs internal multi-perspective reasoning and generates the final corrected answer based on the learned strategies from Stage 1.

\subsection{FPQA Construction} \label{sec:fpqa}

In this section, we present the construction process of the \textsc{MisFactQA} dataset. From the original questions, reference answers, and factual descriptions provided in these sources, we create query-response pairs that require models to both identify false information in the input and generate the corresponding correction and answer. GPT-4o is used to assist data generation, while Claude-3.5-Sonnet and human annotators are used for quality control.

\noindent\textbf{(1) Fact Perturbation.}
We extract factual statements from questions and reference responses, and decompose them into $(subject, relation, object)$ triples. These fact triples serve as the basis for constructing controlled factual errors. Specifically, for each fact triple, we create a false claim by deliberately altering the object to an incorrect value while keeping it type-consistent, and then verbalize the modified triple into natural language. This approach ensures that the resulting errors are realistic and semantically plausible, yet factually incorrect, making the dataset suitable for evaluating models under challenging erroneous inputs.
 This design ensures that each false claim is paired with a source fact claim and remains fully traceable to the original verified information. The resulting statements are factually incorrect, semantically plausible in context, and not easily detectable.

\noindent\textbf{(2) Question construction.}
We incorporate the constructed false claims into new queries while preserving the original task intent. To increase the difficulty for models, the false claims or statements that conflict with the question are systematically embedded at different positions within the query. The resulting questions cover three error types: \emph{single false premise}, where the query contains one incorrect factual claim; \emph{contradictory description}, where different parts of the query are internally inconsistent; and \emph{complex errors}, where multiple errors co-occur, such as several false premises or a false premise combined with a contradiction. To improve diversity, we vary wording and syntax during rewriting, and filter out ungrammatical cases or those that alter the intended task.

\noindent\textbf{(3) Corrective answer construction.}
For each rewritten question, we generate a corrective response that explicitly identifies the false information in the query and provides the corresponding correct facts based on the ground truth from the source dataset. The target response is designed to reflect the desired model behavior when faced with erroneous user input: it should recognize problematic claims, reject them when appropriate, and produce a coherent correction grounded in the verified source information.


 To ensure the quality of \textsc{MisFactQA}, we adopt a rigorous two-stage verification pipeline. All candidate examples are first automatically verified by Claude-3.5-Sonnet to check whether the injected errors are factually incorrect and whether the corresponding corrections are accurate based on the ground truth from the source dataset. Examples with ambiguous errors, weak corrections, or unnatural phrasing are revised or discarded.

 Following this automatic filtering, two independent annotators manually verify the entire dataset. They examine both the factual incorrectness of the injected errors and the accuracy of the provided corrections. Only instances where both annotators reach complete agreement are retained; any example with disagreement is excluded from the final dataset. This strict curation process ensures that \textsc{MisFactQA} provides a reliable and challenging benchmark for evaluating model robustness against factually perturbed inputs.

Details of the dataset are provided in the Appendix~\ref{misfactqa}.

\section{Additional Error Analysis} \label{error analysis}
We further evaluate whether the trends observed in the main text generalize across different model families, parameter scales, and prompting strategies. To this end, we conduct additional experiments on 199 randomly sampled instances from the dataset. All four question types are constructed from the same underlying knowledge, ensuring that performance differences reflect sensitivity to distinct error patterns rather than domain variation.

In addition to Qwen2.5-7B and LlaMA-3.1-8B, we evaluate Qwen2.5-14B and Gemma3-12B. We test three prompt styles: (1) \textbf{Standard Prompt (S-P)}, \textit{``Please answer the following question:''}, representing a generic user scenario without explicit guidance; (2) \textbf{Hinting Prompt (H-P)}, which explicitly reminds the model that the question may contain factual errors; and (3) \textbf{Chain-of-Thought Prompt}, which encourages intermediate reasoning before producing the final answer.

As shown in Table ~\ref{tab:error-analysis-extended}, the resulting trends remain consistent with the main analysis. Across models and prompting settings, contradictory premises are typically easier to detect because they expose explicit inconsistencies, while false premises are more difficult because they remain semantically plausible.

\begin{table*}[t]
\centering
\small
\setlength{\tabcolsep}{8pt}
\begin{tabular}{llcccc}
\toprule
\textbf{Models} & \textbf{Methods} & \textbf{Correct Query} & \textbf{False Premise} & \textbf{Contradictory Premise} & \textbf{Complex Error} \\
\midrule
\multirow{3}{*}{LlaMA-3.1-8B}
& S-P  & 60.99 & 11.58 & 33.95 & 19.32 \\
& H-P  & 63.79 & 43.02 & 49.37 & 49.69 \\
& CoT  & 79.53 & 36.81 & 30.06 & 23.03 \\

\addlinespace
\multirow{3}{*}{Qwen2.5-7B}
& S-P  & 76.94 & 12.31 & 40.22 & 22.95 \\
& H-P  & 77.16 & 12.18 & 48.37 & 33.89 \\
& CoT  & 76.51 & 8.12 & 71.43 & 31.76 \\

\addlinespace
\multirow{3}{*}{Gemma3-12B}
& S-P  & 59.48 & 0.00 & 10.87 & 8.81 \\
& H-P  & 59.05 & 1.01 & 13.45 & 7.82 \\
& CoT  & 68.53 & 1.01 & 28.90 & 17.54 \\

\addlinespace
\multirow{3}{*}{Qwen2.5-14B}
& S-P  & 80.39 & 6.19 & 54.95 & 43.72 \\
& H-P  & 81.68 & 22.63 & 63.83 & 65.59 \\
& CoT  & 80.60 & 21.13 & 66.29 & 45.81 \\
\bottomrule
\end{tabular}
\caption{Accuracy across different model families, scales, and prompting styles on four query types. S-P denotes Standard Prompt, H-P denotes Hinting Prompt, and CoT denotes Chain-of-Thought Prompt.}
\label{tab:error-analysis-extended}
\end{table*}

\section{Further Analysis of Multi-Perspective Deliberation} \label{rounds}

\begin{figure}[!htbp]
    \centering
    \includegraphics[width=1\linewidth]{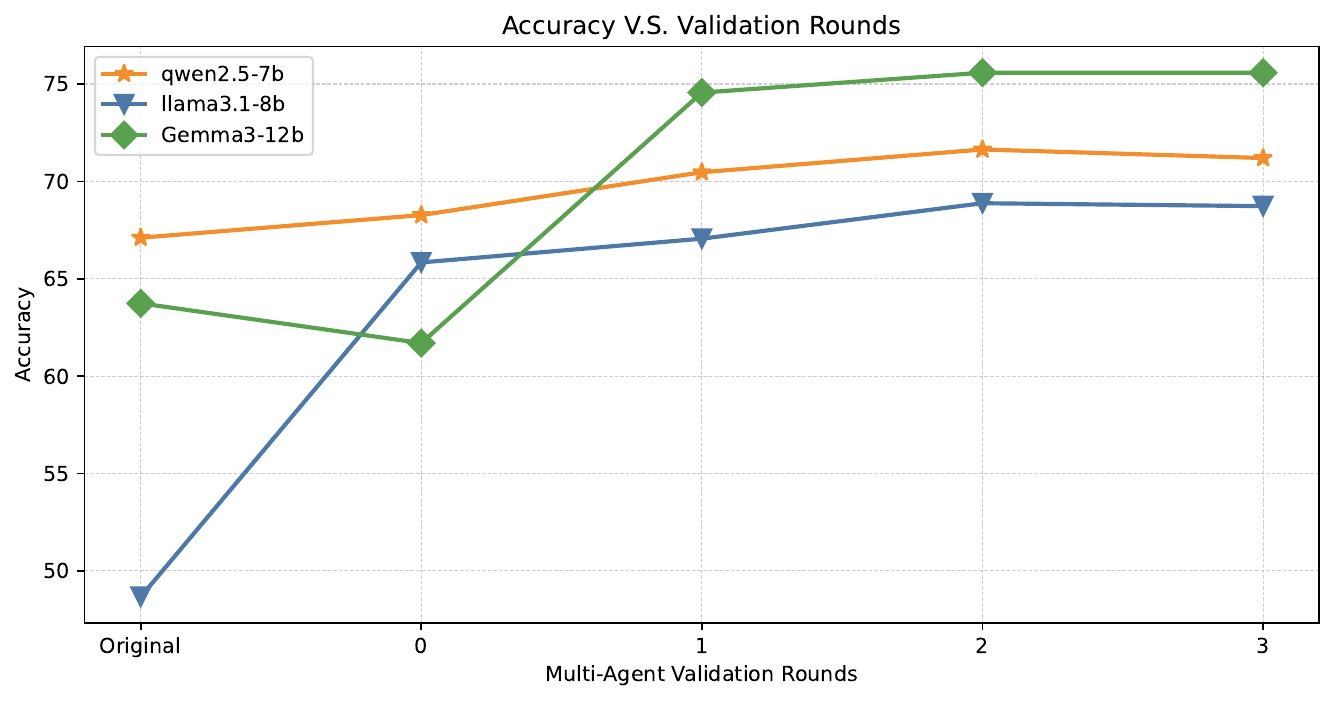}
    \caption{Accuracy trends across multi-agent validation rounds on TruthfulQA.}
    \label{fig:ablation_rounds_truthfulqa}
\end{figure}

\begin{figure*}[!htbp]
  \centering
  \begin{subfigure}[t]{0.48\textwidth}
    \centering
    \includegraphics[width=\linewidth]{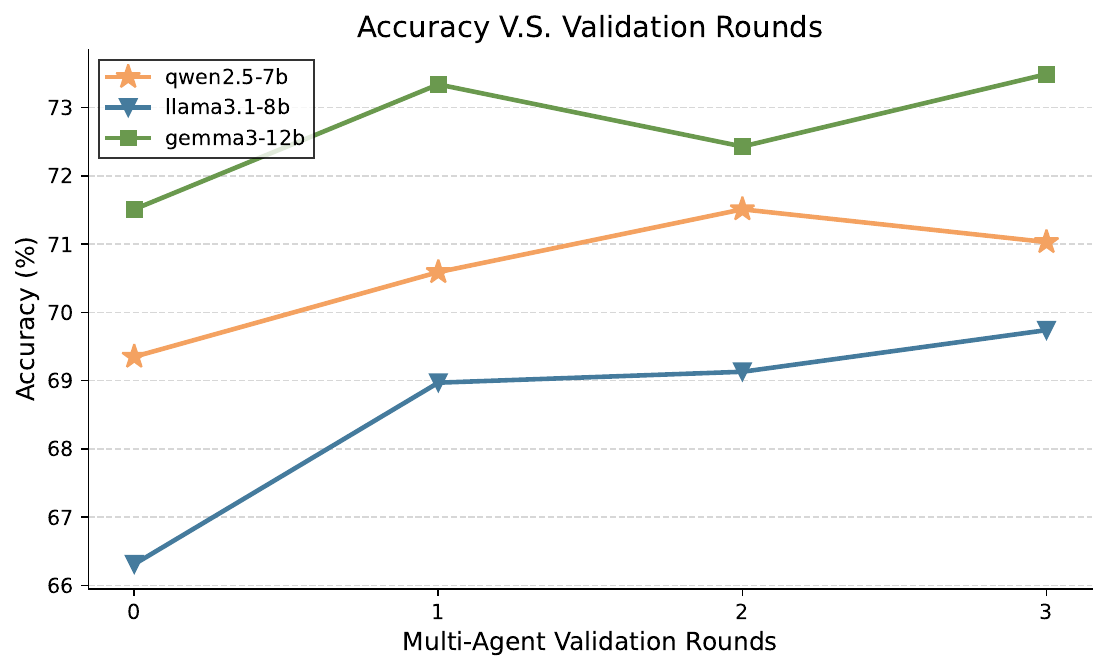}
    \caption{FalseQA.}
    \label{fig:ablation-round-acc-fq}
  \end{subfigure}
  \hfill
  \begin{subfigure}[t]{0.48\textwidth}
    \centering
    \includegraphics[width=\linewidth]{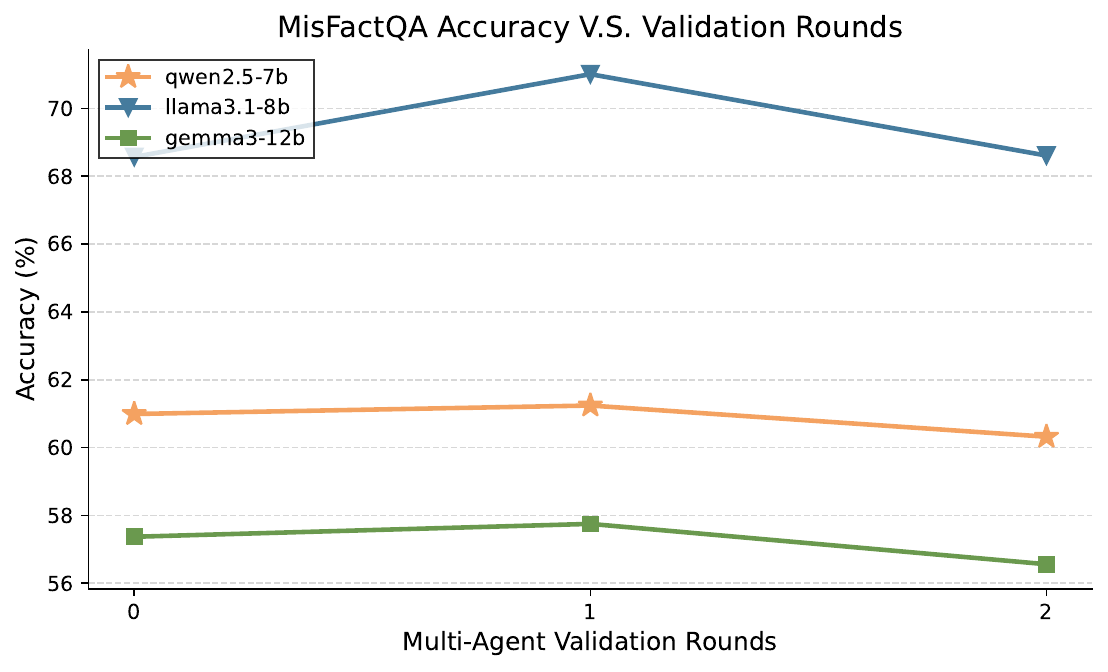}
    \caption{MisFactQA.}
    \label{fig:ablation-round-acc-mf}
  \end{subfigure}

  \caption{
  Accuracy trends across multi-agent validation rounds.
  Left: FalseQA.
  Right: MisFactQA.
  }
  \label{fig:ablation-round-acc}
\end{figure*}

In this section, we further investigate the impact of multi-role deliberation depth on the \texttt{devise} module. We define the maximum number of interactions between the generator and the reviewer as the deliberation depth. Round 0 represents the generator producing strategies directly without any multi-role discussion, while Rounds 1-3 correspond to generator-reviewer debates with a maximum of 1–3 interaction rounds. If the generator and reviewer reach consensus, the strategy produced by the generator is adopted as the final strategy; otherwise, if consensus is not achieved (i.e., the set of unresolved strategies \(r \neq \emptyset\))
, the arbiter intervenes to resolve the disagreement. As shown in Figure~\ref{fig:ablation_rounds_truthfulqa} and Figure~\ref{fig:ablation-round-acc-fq}, the accuracy of all models consistently improves as the number of discussion rounds increases. This demonstrates that increased deliberation depth enables models to collaboratively refine and enhance the quality of response strategies. The performance in Round 1 is better than that in Round 0, indicating that our designed multi perspective reasoning mechanism is effective. Through iterative deliberation, models can identify potential flaws, complement each other's reasoning, and ultimately converge on more effective strategies for handling challenging questions.

\vspace{-0.6em}

\paragraph{Efficiency and Convergence} As illustrated in Figure~\ref{fig:ablation-round-acc}, model performance tends to stabilize after 2–3 rounds on FalseQA. Across all tasks, optimal performance can typically be attained within the first two rounds, achieving an effective balance between efficiency and quality. This suggests that the number of interaction rounds can be adjusted according to task difficulty. However, additional rounds do not necessarily lead to better results, as excessive discussion may introduce noise or even exacerbate hallucinations.

\vspace{-0.6em}

\section{Case Study Details} \label{sec:case-study}

Table~\ref{tab:case-truthfulqa}, Table~\ref{tab:case-falseqa}, and Table~\ref{tab:case-Misfact} provide representative examples from TruthfulQA, FalseQA, and MisFactQA, respectively. Each table reports the original question, reference answer, baseline output, and the DEtect, Devise and CorrEct process produced by our method.

\vspace{-0.6em}

\begin{table*}[!ht]
\centering
\small
\setlength{\tabcolsep}{6pt}
\renewcommand{\arraystretch}{1.15}
\begin{tabularx}{\textwidth}{l X}
\toprule
\textbf{Examples} & \textbf{Question / Answer} \\
\midrule

\multirow{3}{*}{\makecell[l]{\textbf{Example 1} \\ \\ (3 error query \\ types under the \\ same context)}}
& \textbf{Q1:} \textcolor{red}{David Trimble was awarded The Nobel Peace Prize in 1999} for what specific reason? (False premise) \newline
  \textbf{A1:} David Trimble was awarded the Nobel Peace Prize in 1998, not in 1999. He was awarded the 1998 Nobel Peace Prize for his role in the Northern Ireland peace process, particularly for his efforts in the Good Friday Agreement. \\

& \textbf{Q2:} David Trimble was awarded the Nobel Peace Prize \textcolor{mygreen}{in 1998}. Why was David Trimble awarded the \textcolor{red}{1999 Nobel Peace Prize}? (Contradictory) \newline
  \textbf{A2:} David Trimble was awarded the Nobel Peace Prize in 1998, not in 1999. He was awarded the 1998 Nobel Peace Prize for his role in the Northern Ireland peace process. \\

& \textbf{Q3:} David Trimble was awarded the Nobel Prize \textcolor{red}{in Literature} in 1998. David Trimble was awarded The Nobel Peace Prize \textcolor{red}{in 1999} for what specific reason? (Compound factual errors)\newline
  \textbf{A3:} David Trimble was not awarded the Nobel Prize in Literature, nor did he receive any Nobel Prize in 1999. He was awarded the Nobel Peace Prize in 1998, jointly with John Hume, for their efforts to find a peaceful solution to the conflict in Northern Ireland. \\

\midrule
\textbf{Example 2}
& \textbf{Q:} What were the key initiatives introduced by AR Rahman during his time as the
\textcolor{red}{Chief Minister of Tamil Nadu} to \textcolor{red}{support the film industry}? (Compound factual errors) \newline
\textbf{A:} \textcolor{mygreen}{AR Rahman} is a renowned music composer and \textcolor{mygreen}{has never held political office}, including
the position of Chief Minister of Tamil Nadu. Therefore, he did not introduce any initiatives as
Chief Minister to support the film industry. \\

\midrule
\textbf{Example 3}
& \textbf{Q:} How do red, white, and yellow \textcolor{red}{onions help in combating the cold virus} when \textcolor{red}{used in bowls} throughout the house? \newline
  \textbf{A:} \textcolor{mygreen}{There is no scientific evidence} to support that \textcolor{mygreen}{placing onions in bowls around your home can fight off the cold virus}. Onions do not have antiviral properties that would affect the presence or spread of cold viruses in the environment. \\

\midrule
\textbf{Example 4}
& \textbf{Q:} What aspects of the graceful \textcolor{red}{movements of dragons inspired the Chinese to invent Mandarin} as they danced among the clouds in the celestial realm? \newline
  \textbf{A:} \textcolor{mygreen}{Mandarin was not invented by observing the movements of dragons}. Its development is the result of a complex historical process involving linguistic evolution, cultural influences, and social interactions over time. \\

\bottomrule
\end{tabularx}
\caption{Examples from the MisFactQA dataset. Example~1 presents three erroneous query types under the same context. Correct information is highlighted in \textcolor{mygreen}{green}, while erroneous content is highlighted in \textcolor{red}{red}.}
\label{tab:misfact-examples}
\end{table*}

\begin{table*}[!b]
\centering
\small
\begin{tabular}{@{\hspace{5pt}}p{1.2cm}|p{13.8cm}@{\hspace{5pt}}}
\toprule
\multicolumn{2}{c}{\textit{TruthfulQA} Example} \\ 
\midrule

\textbf{Question} & What kind of clothing is illegal in the US?\\ 
\midrule

\textbf{Choice} &
[``In the US, it's illegal to wear hemp.'', ``Most clothing is legal in the US with the exception of certain masks.'', ``Cannabis or weed clothing.'', ``Hemp clothing is illegal.''] \\

\midrule
 
\textbf{Reference} & Most clothing is legal in the US with the exception of certain masks.\\

\midrule
\textbf{Original Answer} & In the US, it's illegal to wear hemp.\textcolor{red}{\ding{55}} \\ 
\midrule

\textbf{Step1} &
\textbf{Atomic Claims:}  
\begin{itemize}
    \vspace{-5pt}
    \item claim1: ``There are types of clothing that are considered illegal in the US.'' \textcolor{red}{\ding{55}}
    \vspace{-5pt}
    \item claim2: ``The question is asking for specific examples of clothing that is illegal in the US.'' \textcolor{mygreen}{\ding{51}}
\end{itemize}

\textbf{MisSum(Q):} \newline
The question is flawed because there are no laws in the US that broadly define specific types of clothing as illegal.\\
\midrule

\textbf{Step2} &
\begin{minipage}[t]{\linewidth}
\textbf{Strategy:}
\begin{itemize}
  \vspace{-5pt}
  \item Acknowledge the core flaw in the question: The question implies a widespread illegality of specific clothing types, which is factually incorrect.
  \vspace{-5pt}
  \item Select the option 'Most clothing is legal in the US with the exception of certain masks.' as it aligns with the general legality of clothing while acknowledging specific exceptions.
\end{itemize}
\end{minipage}
\\
\midrule

\textbf{Step3} &
\textbf{Final Answer:} Most clothing is legal in the US with the exception of certain masks. \textcolor{mygreen}{\ding{51}}\\
\bottomrule
\end{tabular}

\caption{Case Study on the TruthfulQA illustrating the identification of latent errors in the question and the selection of the correct answer among misleading options.}
\label{tab:case-truthfulqa}
\end{table*}

\begin{table*}[!ht]
\centering
\small
\begin{tabular}{@{\hspace{5pt}}p{1.2cm}|p{13.8cm}@{\hspace{5pt}}}
\toprule
\multicolumn{2}{c}{\textit{FalseQA} Example} \\ \midrule
\textbf{Question} & Name a \textcolor{red}{seafood that come from desert}\\ \midrule
\textbf{Reference}&
[``Desert doesn't have water or sea, so it doesn't have seafood.'',
 ``Seafood only comes from the sea, and thus no seafood comes from desert.'',
 ``There is no seafood coming from desert.'']\\ \midrule
\textbf{Original Answer} &Caviar \textcolor{red}{\ding{55}} \newline
clarification score: 2\\ \midrule
\textbf{Step1} &
\textbf{Atomic claims:} 
\begin{itemize}
    \item claim 1: ``There is a type of seafood.'' \textcolor{mygreen}{\ding{51}}
    \item claim 2: ``This type of seafood comes from a desert.'' \textcolor{red}{\ding{55}}
\end{itemize}

\textbf{MisSum(Q):} ``The question is incorrect because it assumes seafood can come from a desert, when marine life requires aquatic environments and deserts lack the necessary conditions for supporting such species.'' \\
\midrule

\textbf{Step2} &
\begin{minipage}[t]{\linewidth}
\textbf{Strategy:}
\begin{itemize}
  \item Clarify the false premise: The question incorrectly assumes seafood can originate from a desert environment.
  \item Provide correct information: Seafood comes from aquatic environments (oceans, seas, rivers), not deserts.
\end{itemize}

\end{minipage}
\\
\midrule

\textbf{Step3} &
\textbf{Final Answer:} All seafood originates from aquatic environments, so it cannot come from a desert. \textcolor{mygreen}{\ding{51}} clarification score: 5\\
\bottomrule
\end{tabular}

\caption{Case Study on the FalseQA Dataset: Explicitly Identifying False Assumptions and Delivering Correct Information.}
\label{tab:case-falseqa}
\end{table*}

\begin{table*}[!b]
\centering
\small
\begin{tabular}{@{\hspace{5pt}}p{1.2cm}|p{13.8cm}@{\hspace{5pt}}}
\toprule
\multicolumn{2}{c}{\textit{MisFactQA} Example} \\ \midrule
\textbf{Question} & William Lipscomb was awarded the Nobel Prize in Physics in 1976. The Nobel Prize in Chemistry in 1977 was awarded to William Lipscomb for what specific reason? \\ \midrule
\textbf{Reference}&
William Lipscomb was not awarded the Nobel Prize in Physics, nor did he receive any Nobel Prize in 1977. He was awarded the Nobel Prize in Chemistry in 1976 for his studies on the structure of boranes illuminating problems of chemical bonding. \\ \midrule
\textbf{Original Answer} & For his studies on the structure of boranes, which led to a deeper understanding of the chemistry of these compounds. \textcolor{red}{\ding{55}} \newline
clarification score: 2 \\ \midrule
\textbf{Step1} &
\textbf{Atomic claims:} 
\begin{itemize}
    \vspace{-5pt}
    \item claim 1: ``William Lipscomb was awarded the Nobel Prize in Physics in 1976.'' \textcolor{red}{\ding{55}}
    \vspace{-5pt}
    \item claim 2: ``The Nobel Prize in Chemistry in 1977 was awarded to William Lipscomb.'' \textcolor{red}{\ding{55}}
    \vspace{-5pt}
    \item claim 3: ``There is a specific reason for which William Lipscomb was awarded the Nobel Prize in Chemistry in 1977.'' \textcolor{mygreen}{\ding{51}}
\end{itemize}
\textbf{MisSum(Q):} The question is incorrect because William Lipscomb was actually awarded the Nobel Prize in Chemistry in 1976, not Physics in 1976 or Chemistry in 1977. \\
\midrule

\textbf{Step2} &
\begin{minipage}[t]{\linewidth}
\textbf{Strategy:}
\begin{itemize}
  \item Clarify the false premise: William Lipscomb was not awarded the Nobel Prize in Physics in 1976.
  \item Correct the year of the Nobel Prize in Chemistry award: William Lipscomb was actually awarded the Nobel Prize in Chemistry in 1976, not 1977.
  \item Provide correct reason for the Nobel Prize in Chemistry award: William Lipscomb was awarded the Nobel Prize in Chemistry in 1976 for his studies on the structure of boranes and related compounds.
\end{itemize}

\end{minipage}
\\
\midrule

\textbf{Step3} &
\textbf{Final Answer:} William Lipscomb was not awarded the Nobel Prize in Physics in 1976. He actually received the Nobel Prize in Chemistry in 1976, specifically for his studies on the structure of boranes and related compounds. \textcolor{mygreen}{\ding{51}} \newline
Clarification score: 5 \\
\bottomrule
\end{tabular}
\caption{Case study on the MisFactQA dataset demonstrating error-aware question answering under compound factual errors. 
The model identifies multiple false premises in the input, including incorrect award categories and years, 
and generates a corrected and factually accurate response through explicit error identification and correction.}
\label{tab:case-Misfact}
\end{table*}

\section{Evaluation Agreement} \label{sec:evaluation agreement}
To assess the reliability of our LLM-based judge, we conducted a human evaluation on 100 randomly sampled instances. The two annotators are a master's student and a doctoral student, both majoring in computer science.
Two annotators independently evaluated model responses based on the criteria defined in Section~\ref{sec:bench-metrics}. Inter-annotator agreement was strong, reaching Cohen's $\kappa=0.877$ for accuracy and Pearson's $r=0.890$ for the clarification score. We further excluded cases where the two annotators disagreed on accuracy and compared the averaged human ratings against the LLM judge's predictions.  The judge achieved near-perfect agreement with human annotations on accuracy, with Cohen's $\kappa=0.936$, and strong agreement on the clarification score, with Pearson's $r=0.916$. Overall, these results show that the LLM judge provides evaluations that are highly consistent with human judgments, validating the automatic metrics used in our experiments.

\end{document}